\documentclass[11pt]{article}

\usepackage[T1]{fontenc}
\usepackage[utf8]{inputenc}
\usepackage{lmodern}
\usepackage{microtype}
\usepackage[a4paper,margin=1in]{geometry}
\usepackage{amsmath,amssymb}
\usepackage{booktabs}
\usepackage{tabularx}
\usepackage{array}
\usepackage{graphicx}
\usepackage{caption}
\usepackage{subcaption}
\usepackage{enumitem}
\usepackage{xcolor}
\usepackage{hyperref}
\usepackage{url}
\usepackage{float}

\hypersetup{
  colorlinks=true,
  linkcolor=blue!45!black,
  citecolor=blue!45!black,
  urlcolor=blue!45!black,
  pdftitle={QuantCode Model: Specializing Language Models for Executable Algorithmic Trading Code},
  pdfauthor={Alexey Chernysh, Orkhan Ekhtibarov, Dmitry Zmitrovich}
}
\setlist{nosep}

\title{\textbf{QuantCode Model: Specializing Language Models for Executable Algorithmic Trading Code}}

\author{
Alexey Chernysh\thanks{Corresponding author: \href{mailto:aleksei.0.chernysh@gmail.com}{\texttt{aleksei.0.chernysh@gmail.com}}} \quad
Orkhan Ekhtibarov \quad
Dmitry Zmitrovich
}
\date{September 2026}

\begin{document}
\maketitle

\begin{abstract}
Large language models are strong general-purpose code generators, but executable algorithmic trading remains a demanding specialization target: a model must translate a natural-language strategy specification into correct program logic for a specialized trading framework, execute on historical data, produce trades, and remain semantically faithful to the request. We study two complementary mechanisms for specializing language models for this setting: continued pretraining on algorithmic-trading framework code and supervised fine-tuning (SFT) on agent-validated request-to-code pairs. Evaluation is centered on QuantCode-Bench, our 400-task benchmark for Backtrader strategy generation, together with a repository-level SWE-bench-like track. Continued pretraining improves single-turn Judge Pass from 41.5\% to 47.5\% for Qwen3.5-397B-A17B and from 27.8\% to 33.0\% for Qwen3.6-35B-A3B. SFT applied after continued pretraining yields a larger gain for Qwen3.6-35B-A3B, reaching 58.2\% Judge Pass and 83.5\% successful backtests; in agentic evaluation it raises first-turn success from 22.3\% to 58.3\% and final success after up to 10 turns from 47.5\% to 79.5\%. Continued pretraining alone improves first-turn agentic success but lowers final success after repair from 47.5\% to 32.5\%, consistent with degraded instruction following, whereas SFT improves both. We also identify a capability-retention failure: domain specialization degrades parser-conformant structured tool calling, and targeted recovery SFT restores tool-call formatting but not the base checkpoint's repository-level agent performance. The results show that framework-oriented pretraining, validated SFT, and explicit capability-retention evaluation address distinct failure modes in domain-specific executable code generation.
\end{abstract}

\section{Introduction}

Large language models (LLMs) are increasingly used as software-generation systems rather than as purely conversational assistants. This shift changes what it means for a model output to be correct. In natural-language generation, a response can often be evaluated directly as text. In software generation, the response is an executable artifact whose validity depends on syntax, dependencies, framework APIs, runtime behavior, state transitions, and alignment with the user's intended program. These requirements become more stringent in specialized software domains, where a model can be proficient in general Python programming yet still fail systematically because it does not know the conventions, abstractions, or execution semantics of a particular framework.

Algorithmic trading is an especially demanding instance of this problem. A natural-language request may specify a combination of indicators, entry conditions, exit conditions, position sizing, stop logic, time constraints, and risk-management rules. Translating such a request into a functioning strategy requires the model to identify the intended financial logic, map it to a target framework such as Backtrader \cite{backtrader}, construct indicators with the correct data dependencies, manage state across bars, submit and track orders, and avoid implementation mistakes that only become visible at runtime. A program may be syntactically correct yet fail to initialize. It may execute successfully but never satisfy its own entry conditions. It may generate trades but implement a different strategy from the one requested. These failure modes cannot be captured reliably by text similarity or compilation accuracy alone.

Our earlier work introduced QuantCode-Bench to make this distinction measurable \cite{khoroshilov2026quantcode}. The benchmark contains 400 textual descriptions of algorithmic trading strategies and evaluates generated Backtrader code through a staged executable pipeline. A candidate must be syntactically valid, execute successfully in the backtest environment, produce at least one trade, and pass an LLM-based semantic judge that verifies consistency with the requested strategy. The benchmark supports both a single-turn setting and an agentic multi-turn setting in which a model receives structured tool feedback and may revise its code for up to 10 attempts. The benchmark therefore decomposes a seemingly simple text-to-code problem into several nested capabilities: code generation, framework use, executable strategy construction, production of trading behavior, and semantic alignment.

The present study asks a different question from the benchmark paper. Rather than comparing general-purpose models, we investigate whether targeted training can alter the capability profile measured by this execution-centered benchmark. We consider two distinct mechanisms. First, continued pretraining can expose a base model to the source code and usage patterns of algorithmic-trading frameworks. This intervention targets domain and API familiarity. Second, SFT can provide a direct mapping from realistic user requests to validated framework-conformant strategy implementations. This intervention targets instruction-to-code alignment.

These two mechanisms are related but not interchangeable. Continued pretraining teaches what domain code looks like but does not directly optimize a user-request mapping. SFT teaches such a mapping, and in our pipeline its targets are validated in the execution environment, but each demonstration remains a static target rather than a trajectory of iterative repair. Consequently, evaluating both stages within the same model lineage, in both single-turn and agentic settings, provides a more complete picture of specialization than a study of either stage alone.

The experiments form directly comparable sequential training lineages. For Qwen3.6-35B-A3B, training proceeds from the base checkpoint through continued domain pretraining and then SFT, so the base, pretraining, and SFT measurements can be interpreted as successive stages of the same 35B lineage. For Qwen3.5-397B-A17B, the domain-pretrained checkpoint is likewise compared directly with its own base checkpoint. We therefore compare pretraining and SFT directly within their sequential lineages.

The results show that the two stages act on different parts of the pipeline. Continued pretraining yields moderate but consistent gains in single-turn Judge Pass. SFT yields a much larger improvement on the Qwen3.6-35B-A3B checkpoint, raising not only the semantic success rate but also successful backtesting and trade generation. Agentic evaluation shows that the SFT model begins from a much stronger first attempt and remains substantially better after iterative repair. Continued pretraining alone, by contrast, raises first-turn agentic success but lowers success after iterative repair below the base checkpoint, which we interpret as degraded processing of conversational feedback.

We additionally extend evaluation beyond isolated strategy files to a repository-level software-engineering track inspired by SWE-bench \cite{jimenez2024swebench}. These tasks begin from real algorithmic-trading repositories and require the model to inspect repository context, respond to an issue, generate a patch, run repository tests, and repair the code from tool feedback. This setting probes whether the same domain-specific capabilities transfer from self-contained strategy generation to long-horizon software engineering.

A separate deployment-oriented experiment reveals an important capability-retention failure. The same domain specialization that improves QuantCode-Bench can degrade structured tool-call formatting required by an interactive agent harness. We therefore evaluate two recovery strategies---weight merging and a dedicated tool-use SFT stage---and distinguish restoration of parser-conformant calls from restoration of downstream tool-use quality. This analysis highlights a capability that the current QuantCode-Bench protocol does not directly measure.

The contribution of this report is therefore a training-centered complement to QuantCode-Bench. We provide a unified description of the data construction, objectives, target validation, and evaluation results for continued pretraining and SFT. We also show why the notion of success in this domain must remain behaviorally grounded: generating code that looks plausible is not sufficient if the code does not execute, does not trade, or does not implement the requested strategy.

\section{Related Work}

\paragraph{Domain-adaptive and code-centric pretraining.}
Domain-adaptive pretraining has long been used to specialize general language models toward target distributions \cite{gururangan2020dontstop}. In code modeling, Code Llama \cite{roziere2023codellama}, StarCoder2 \cite{lozhkov2024starcoder2}, and DeepSeek-Coder \cite{guo2024deepseekcoder} demonstrate that corpus composition and code-centric pretraining materially affect programming capability. These systems specialize at the programming-language or repository level across broad code distributions. Our continued-pretraining experiments narrow the target distribution further, focusing on source code and usage patterns associated with algorithmic-trading frameworks and repositories. The objective is not merely stronger general coding performance, but improved familiarity with the APIs, lifecycle methods, indicators, order handling, and framework conventions required by executable trading strategies.

\paragraph{General-purpose model families and technical reports.}
The experiments use Qwen-family checkpoints. The Qwen3 technical report describes dense and mixture-of-experts models spanning multiple parameter scales and emphasizes coding, reasoning, and agentic capabilities \cite{yang2025qwen3}. We use these models as starting points rather than proposing a new foundation architecture. Recent technical reports such as GLM-5 also emphasize a shift from static coding to long-horizon agentic engineering \cite{zeng2026glm5}. Our work follows this behavioral perspective but evaluates it in a narrowly defined, executable financial software domain.

\paragraph{Repository-level software engineering.}
SWE-bench established real GitHub issue resolution as a benchmark for language-model software engineering \cite{jimenez2024swebench}. Its central insight is that repository-level tasks require more than isolated function generation: models must understand existing code, identify relevant files, generate patches, and interact with executable tests. We adopt the same general paradigm for an algorithmic-trading repository track. The distinction is the domain of the repositories, issues, and source code under modification; verification itself follows an executable test-based protocol analogous to SWE-bench.

\paragraph{Financial-domain evaluation.}
Our separate FINESSE-Bench work evaluates financial-domain knowledge and technical-analysis reasoning \cite{stanishevskii2026finesse}. The present report deliberately separates financial knowledge from executable software competence. A model may know technical-analysis concepts while still failing to implement a Backtrader strategy correctly; conversely, a model may reproduce framework patterns without possessing broad financial expertise. QuantCode-Bench and the training experiments in this report target the latter capability: faithful operationalization of a trading request into executable code.

\section{Methods}

\subsection{Task formulation}

Given a natural-language strategy request \(x\), the model generates program \(y\) in a target algorithmic-trading framework. Correctness is treated as a sequence of increasingly restrictive executable conditions. We define
\[
B(y)\in\{0,1\}
\]
to indicate successful execution in the backtesting environment,
\[
T(y)\in\{0,1\}
\]
to indicate that the strategy produces at least one trade, and
\[
J(x,y)\in\{0,1\}
\]
to indicate that an LLM judge determines that the generated implementation is semantically aligned with the requested strategy.

The strict terminal success criterion is
\[
S(x,y)=\mathbb{1}\left[B(y)=1 \land T(y)=1 \land J(x,y)=1\right].
\]
QuantCode-Bench additionally tracks compilation before execution, but we report the executable conditions above because a strategy that cannot run cannot satisfy \(B(y)\). The benchmark methodology and the role of semantic judging are described in detail in \cite{khoroshilov2026quantcode}.

The objective is implementation correctness rather than profitability. A strategy can correctly implement a user's idea and still lose money over a particular historical window. Including return, Sharpe ratio, or another economic target in the success criterion or in the selection of training targets would change the problem from code generation to strategy discovery and could favor deviations from the user's request. We therefore restrict the evaluation signal and the validation of training targets to technical executability, trading behavior, and semantic compliance.

\subsection{Experimental structure and comparability}

The report covers two training regimes, one additional repository-level evaluation, and a tool-call recovery intervention (Sections~\ref{sec:toolrecovery-method} and~\ref{sec:toolrecovery-results}). Table~\ref{tab:regimes} summarizes their roles. The main QuantCode-Bench training results contain two directly comparable lineages: Qwen3.6-35B-A3B follows \texttt{base -> continued pretraining -> SFT}, while Qwen3.5-397B-A17B follows \texttt{base -> continued pretraining}.

\begin{table}[H]
\centering
\caption{Training and evaluation regimes.}
\label{tab:regimes}
\begin{tabularx}{\textwidth}{@{}lX X X@{}}
\toprule
Regime & Starting point & Training signal & Primary evaluation \\
\midrule
Continued pretraining & Qwen3.6-35B-A3B and Qwen3.5-397B-A17B checkpoints & Algorithmic-trading framework/repository code & Current QuantCode-Bench \\
SFT & Qwen3.6-35B-A3B domain-pretrained checkpoint & Validated user request \(\rightarrow\) Backtrader code & Current QuantCode-Bench \\
Repository track & General-purpose models (specialized checkpoints not evaluable, Section~\ref{sec:toolrecovery-results}) & Evaluation only & SWE-bench-like algorithmic-trading tasks \\
Tool-call recovery & Linear merge of the Qwen3.6-35B-A3B SFT checkpoint with the original (0.65/0.35) & Weight merging; SFT on Qwen3.5-122B-A10B agentic trajectories & Parser-conformance checks and SWE-bench-like track, reasoning disabled \\
\bottomrule
\end{tabularx}
\end{table}

This structure permits direct stage-by-stage comparison for the larger-model lineages. In the Qwen3.6-35B-A3B sequence, the effect of continued pretraining can be measured against the base checkpoint and the additional effect of SFT can be measured against the domain-pretrained checkpoint, as well as cumulatively against the base. The Qwen3.5-397B-A17B pretraining effect is likewise a direct base-to-pretraining comparison.

\subsection{Continued pretraining on algorithmic-trading code}

The continued-pretraining stage targets framework and repository familiarity. The principal code component consists of GitHub repositories associated with algorithmic trading and, in particular, source code exposing real usage patterns of trading frameworks. This component was collected by querying GitHub with a set of keywords describing trading and backtesting (framework names, strategy and indicator vocabulary, market-data and broker terminology), downloading the matched repositories, and retaining only those that pass popularity and recency filters (minimum star count and a most-recent-commit date threshold). The retained corpus was then deduplicated with MinHash-based near-duplicate detection to remove forks, vendored copies, and near-identical strategy templates that are common in this domain. The resulting corpus contains approximately 5--6B tokens. The target distribution includes strategy lifecycle methods, indicators, data feeds, order submission and tracking, position state, broker interactions, and the surrounding abstractions used by framework-native code.

This stage differs from SFT because there is no explicit user-request/answer mapping. The model is trained to continue domain code sequences, increasing the probability that relevant APIs and programming idioms are represented in its parameters before instruction alignment. The intervention is motivated by the same broad principle as domain-adaptive pretraining \cite{gururangan2020dontstop}, but the domain here is a software ecosystem rather than a natural-language subject area.

We evaluate continued pretraining at two scales. One experiment compares the Qwen3.5-397B-A17B checkpoint with its algorithmic-trading GitHub-pretrained counterpart. A second compares the Qwen3.6-35B-A3B checkpoint with a domain-pretrained counterpart. Both continued pretraining and the subsequent SFT stage update all model parameters (full fine-tuning) in bf16, using FSDP2 with 8-way expert parallelism. Continued pretraining uses AdamW (\(\beta_1=0.9\), \(\beta_2=0.95\), no weight decay) with a peak learning rate of \(1\times10^{-4}\) and cosine decay to \(1\times10^{-6}\), a global batch size of 512 sequences of 8192 tokens, and 2{,}250 optimizer steps. Because these were internal training snapshots, this report focuses on the observed downstream effect rather than presenting a new general pretraining recipe.

\subsection{Supervised fine-tuning data}

SFT is constructed as a single-state user-request-to-code task. Each input is a natural-language description of a desired trading strategy. Each output is a complete Backtrader implementation that has passed a validation process.

Strategy requests are derived from strategy-related material collected from Reddit, StackExchange, and GitHub, so that the request distribution approximates realistic user intent rather than templated synthetic prompts. Collected items are ranked by a model-based judge for quality and relevance to the target task. When an original description contains useful trading logic but is not phrased as an actionable request, it is rewritten into a realistic user query while preserving the underlying strategy logic; this converts fragmented forum explanations, code-context comments, and repository descriptions into prompts suitable for a code-generation model. QuantCode-Bench tasks were excluded from the request pool when the SFT dataset was assembled.

The resulting SFT dataset contains approximately 800 validated request-to-code pairs. SFT uses AdamW (\(\beta_1=0.9\), \(\beta_2=0.95\), weight decay 0.1) with a constant learning rate of \(1\times10^{-5}\), a global batch size of 128, and 3 epochs, without sequence packing.

A key design choice is that the code targets are generated and checked in an agentic environment where the framework context is available. The agent is shown the Backtrader framework and is encouraged to use the framework's native abstractions rather than reimplementing generic infrastructure from scratch. This affects target quality in two ways. First, it biases examples toward idiomatic framework use: native indicators, lifecycle callbacks, broker methods, order objects, and data lines. Second, it makes it easier to reject targets that are superficially plausible but incompatible with the runtime.

Candidate responses are validated before inclusion. Validation detects malformed imports, invalid API calls, initialization failures, runtime exceptions, incorrect access to framework objects, and other execution defects. The SFT objective is the standard conditional negative log-likelihood
\[
\mathcal{L}_{\mathrm{SFT}}(\theta)
=
-\sum_{t=1}^{|y|}
\log p_\theta(y_t \mid x, y_{<t}),
\]
so the methodological novelty lies primarily in the construction and executable validation of the target distribution rather than in the supervised objective itself.

The SFT stage is applied after continued domain pretraining in the Qwen3.6-35B-A3B lineage and is evaluated under the same current QuantCode-Bench protocol as the corresponding base and domain-pretrained checkpoints. The three 35B measurements therefore form a directly comparable sequential \texttt{base -> continued pretraining -> SFT} progression.

\subsection{QuantCode-Bench evaluation}

QuantCode-Bench contains 400 algorithmic-trading strategy-generation tasks \cite{khoroshilov2026quantcode}. In the single-turn setting, the model receives one opportunity to produce a strategy. We report three downstream metrics from the current experiment snapshot:
\begin{itemize}
    \item \textbf{Backtest}: the strategy executes successfully in the benchmark backtesting environment;
    \item \textbf{Trades}: the strategy produces at least one trade;
    \item \textbf{Judge}: the implementation passes the final semantic-alignment judge.
\end{itemize}

All QuantCode-Bench generations in this report use greedy decoding (temperature 0), and the semantic judge is GPT-5.4. Compilation is also part of the benchmark pipeline, but the training comparison in this report focuses on the stages where specialized models differ most strongly.

In the agentic setting, the model receives structured feedback after a failed attempt and may retry for up to 10 turns. We denote cumulative judge success after different turn budgets by \(T1\), \(T3\), \(T5\), and \(T10\). For the principal Qwen3.6-35B-A3B checkpoint comparison reported below, only \(T1\) and \(T10\) are used; intermediate turn budgets are omitted rather than displayed as missing values.

\subsection{Repository-level algorithmic-trading track}

To evaluate software-engineering behavior beyond standalone strategy generation, we construct a SWE-bench-like track from algorithmic-trading GitHub repositories. The workflow begins with repositories, issues, pull requests, and associated patches. Candidate tasks are packaged into executable environments using repository-specific tooling and are filtered to retain cases with reproducible automated test-based verification.

The agent loop consists of four broad operations: reading the issue and repository context, editing the codebase, running the repository tests, and repairing the patch using tool feedback. The track contains approximately 200 verified tasks in the current snapshot. Because the track requires parser-conformant tool calls, the domain-specialized checkpoints could not be evaluated on it directly (Section~\ref{sec:toolrecovery-results}). The reported metric is the percentage of tasks solved under the corresponding agent configuration.

The defining difference from generic repository benchmarks is the task distribution rather than the acceptance mechanism: the repositories, issues, and code under modification are specific to algorithmic trading, while success is determined by the packaged repository tests in the same general manner as SWE-bench \cite{jimenez2024swebench}. Backtrader backtests are not executed as part of this repository-level benchmark.

\subsection{Tool-call recovery fine-tuning}
\label{sec:toolrecovery-method}

The intended deployment target for the specialized Qwen3.6-35B-A3B model is an interactive developer plugin, where the model operates as an agent and must emit structured tool calls that are parsed by the serving engine. During pre-deployment testing we observed that both domain-specialized Qwen3.6-35B-A3B checkpoints frequently emitted syntactically malformed tool invocations; the phenomenon and its consequences are reported in Section~\ref{sec:toolrecovery-results}. Because the current QuantCode-Bench protocol does not directly measure parser-conformant tool-call generation, we treat mitigation of this failure mode as a separate intervention with its own evaluation protocol.

Two recovery interventions were attempted. The first is parameter-space weight merging between the specialized checkpoint and the original instruction-tuned checkpoint. Prior work on model soups, task arithmetic, and model merging shows that parameter-space combinations can combine or recover capabilities across related checkpoints \cite{wortsman2022soups,ilharco2023task,yu2024dare}. We therefore use merging as a low-cost recovery baseline, without assuming that a simple linear combination must recover the lost formatting behavior. Concretely, the Qwen3.6-35B-A3B SFT checkpoint was merged with the original Qwen3.6-35B-A3B checkpoint using the linear merge method of MergeKit \cite{goddard2024mergekit}, with weight 0.65 on the specialized checkpoint and 0.35 on the original, i.e.\ \(\theta_{\mathrm{merged}} = 0.65\,\theta_{\mathrm{SFT}} + 0.35\,\theta_{\mathrm{orig}}\) applied uniformly to all parameters.

The second intervention is a dedicated recovery SFT stage on agentic tool-use trajectories. Rather than starting from the unmerged SFT checkpoint, this stage is initialized from the merged checkpoint described above. The rationale is that the merged weights lie closer to the original instruction-tuned model in parameter space, and we hypothesized that the residual trace of the original tool-use behavior would make recovery from this initialization more effective than from the fully specialized checkpoint; this choice was not ablated. The training mixture consists of two equally weighted components: (i) software-engineering trajectories collected with the OpenHands agent framework \cite{wang2025openhands} using Qwen3.5-122B-A10B as the trajectory-generating policy, and (ii) trajectories produced by the same Qwen3.5-122B-A10B policy on our algorithmic-strategy dataset, in both single-turn and multi-turn interaction modes. All trajectories use the native structured tool-call format expected by the serving engine, so the mixture directly supervises well-formed call syntax in both general software-engineering and domain-specific contexts. Each trajectory is trained as a single multi-turn sequence with the standard assistant-only loss mask: the cross-entropy loss is computed on the tokens of every assistant turn in the trajectory, including all intermediate turns that emit tool calls, while system, user, and tool-output tokens serve as context only and receive no loss. Supervising every assistant turn rather than only the final one is the common convention for agentic trajectory SFT \cite{pan2024swegym}, and it is what makes the intermediate tool-call syntax a direct training target. Two variants were trained, differing in the maximum sequence length used for trajectory packing and truncation: 16k and 32k tokens.

Recovery checkpoints are evaluated on the repository-level SWE-bench-like track described above, because repository resolution requires sustained tool use over long horizons and therefore measures downstream tool-use effectiveness. We additionally perform deployment-oriented parser-conformance checks to verify whether the checkpoints can again emit tool calls accepted by the serving engine. A standalone quantitative parse-success rate was not retained for this internal snapshot, so we treat formatting recovery as a qualitative capability check and use repository task resolution as the quantitative downstream metric. All recovery-experiment measurements are performed with reasoning disabled.

\section{Results}

\subsection{Continued pretraining improves single-turn Judge Pass}

Table~\ref{tab:singletraining} compares the current Qwen3.6-35B-A3B and Qwen3.5-397B-A17B training checkpoints. Continued pretraining produces a positive Judge Pass change at both scales. For the Qwen3.5-397B-A17B checkpoint, Judge Pass increases from 41.5\% to 47.5\%, an absolute gain of 6.0 percentage points. Backtest success rises from 65.5\% to 71.5\%, and trade generation rises from 42.8\% to 47.5\%.

For the Qwen3.6-35B-A3B checkpoint, domain pretraining increases Judge Pass from 27.8\% to 33.0\%, a gain of 5.2 points. Backtest success rises from 46.0\% to 53.8\%, and the share of tasks producing at least one trade increases from 28.5\% to 36.8\%.

\begin{table}[H]
\centering
\caption{Single-turn QuantCode-Bench results for the principal training interventions. Values are percentages. The Qwen3.6-35B-A3B rows form a sequential \texttt{base -> continued pretraining -> SFT} lineage. The Qwen3.5-397B-A17B rows form a direct \texttt{base -> continued pretraining} comparison.}
\label{tab:singletraining}
\begin{tabular}{@{}lrrr@{}}
\toprule
Checkpoint & Backtest & Trades & Judge \\
\midrule
Qwen3.5-397B-A17B base & 65.5 & 42.8 & 41.5 \\
Qwen3.5-397B-A17B + algotrading-code pretraining & 71.5 & 47.5 & 47.5 \\
\midrule
Qwen3.6-35B-A3B base & 46.0 & 28.5 & 27.8 \\
Qwen3.6-35B-A3B + domain pretraining & 53.8 & 36.8 & 33.0 \\
Qwen3.6-35B-A3B + domain pretraining + SFT & \textbf{83.5} & \textbf{60.0} & \textbf{58.2} \\
\bottomrule
\end{tabular}
\end{table}

\begin{figure}[H]
\centering
\begin{subfigure}{0.49\textwidth}
\centering
\includegraphics[width=\linewidth]{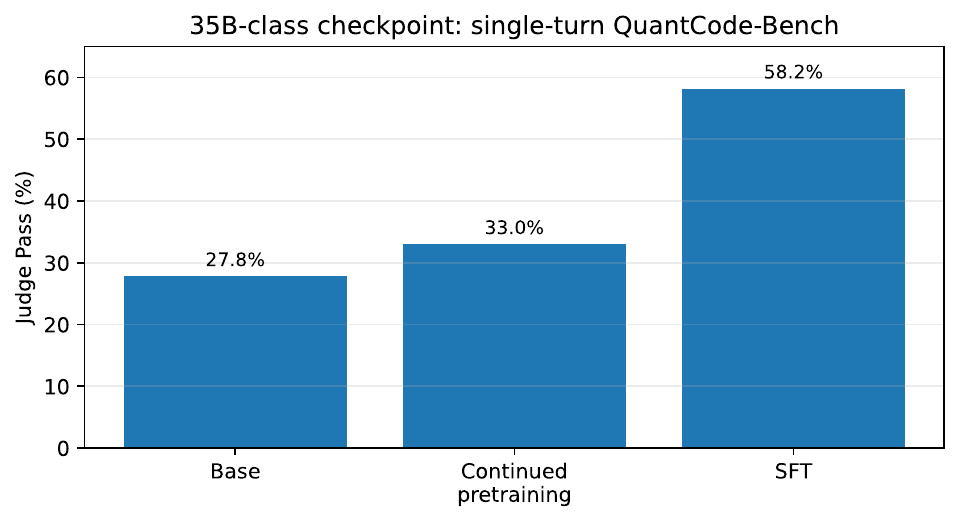}
\caption{Qwen3.6-35B-A3B checkpoint.}
\end{subfigure}
\hfill
\begin{subfigure}{0.43\textwidth}
\centering
\includegraphics[width=\linewidth]{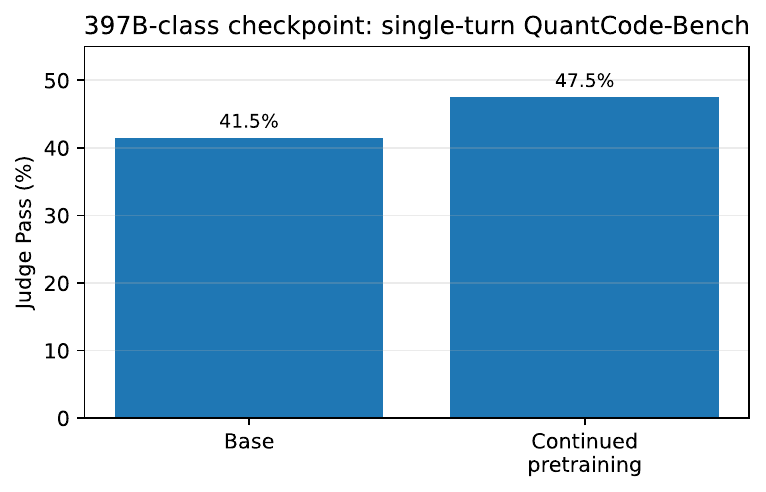}
\caption{Qwen3.5-397B-A17B checkpoint.}
\end{subfigure}
\caption{Single-turn Judge Pass under domain specialization.}
\label{fig:single}
\end{figure}

Figure~\ref{fig:single} visualizes the Judge Pass changes. The magnitude of the continued-pretraining gains is moderate, whereas the SFT gain on the Qwen3.6-35B-A3B model is substantially larger.

\subsection{SFT produces the largest observed single-turn improvement}

The final Qwen3.6-35B-A3B checkpoint after domain pretraining and SFT reaches 83.5\% Backtest, 60.0\% Trades, and 58.2\% Judge Pass. Because SFT is the next stage of the same 35B lineage, its incremental contribution can be measured directly against the domain-pretrained checkpoint: Backtest rises from 53.8\% to 83.5\% (+29.7 points), Trades from 36.8\% to 60.0\% (+23.2 points), and Judge Pass from 33.0\% to 58.2\% (+25.2 points). Cumulatively relative to the original Qwen3.6-35B-A3B base checkpoint, the corresponding gains are 37.5, 31.5, and 30.4 points.

The improvement is therefore not confined to the final LLM judge. The largest change occurs at the executable backtest stage, indicating that the validated request-to-code distribution substantially changes the probability that the model emits runnable framework-conformant strategies. The trade-generation increase shows that more of these runnable programs also reach meaningful trading actions, while the Judge Pass increase indicates that a large fraction preserve the requested semantics.

For context, Table~\ref{tab:leaderboard} shows selected entries from the current single-turn leaderboard snapshot. The specialized Qwen3.6-35B-A3B SFT checkpoint remains below the strongest frontier models but is substantially stronger than its unspecialized base checkpoint and competitive with much larger general-purpose systems on this domain-specific task.

\begin{table}[H]
\centering
\caption{Selected current single-turn QuantCode-Bench results. The table is included for context rather than as a complete leaderboard.}
\label{tab:leaderboard}
\begin{tabular}{@{}lrrr@{}}
\toprule
Model & Backtest & Trades & Judge \\
\midrule
Claude Opus 4.6 & 98.2 & 77.2 & 75.8 \\
GPT-5.4 & 95.5 & 72.0 & 70.2 \\
Claude Sonnet 4.5 & 91.5 & 71.2 & 69.8 \\
GPT-5.2-Codex & 94.5 & 74.5 & 67.5 \\
GLM-5 & 92.4 & 70.3 & 65.4 \\
Gemini-3-Flash & 76.0 & 63.2 & 59.8 \\
Qwen3.6-35B-A3B + pretraining + SFT & 83.5 & 60.0 & 58.2 \\
Qwen3.6-35B-A3B base checkpoint & 46.0 & 28.5 & 27.8 \\
\bottomrule
\end{tabular}
\end{table}

\subsection{SFT strongly improves first-turn and final agentic success}

The agentic results reveal a second effect of the sequential 35B training pipeline. Table~\ref{tab:agentic} reports the available \(T1\) and \(T10\) values for the Qwen3.6-35B-A3B base, domain-pretrained, and subsequent SFT checkpoints. Relative to the domain-pretrained checkpoint, SFT raises first-turn Judge Pass from 32.3\% to 58.3\% (+26.0 points) and \(T10\) success from 32.5\% to 79.5\% (+47.0 points). Relative to the original base checkpoint, the cumulative improvement is from 22.3\% to 58.3\% at \(T1\) (+36.0 points) and from 47.5\% to 79.5\% at \(T10\) (+32.0 points).

\begin{table}[H]
\centering
\caption{Agentic QuantCode-Bench Judge Pass for Qwen3.6-35B-A3B checkpoints at the retained turn budgets.}
\label{tab:agentic}
\begin{tabular}{@{}lrr@{}}
\toprule
Checkpoint & T1 & T10 \\
\midrule
Qwen3.6-35B-A3B base & 22.3 & 47.5 \\
Qwen3.6-35B-A3B domain pretraining & 32.3 & 32.5 \\
Qwen3.6-35B-A3B domain pretraining + SFT & \textbf{58.3} & \textbf{79.5} \\
\bottomrule
\end{tabular}
\end{table}

The domain-pretrained checkpoint shows a higher \(T1\) score than the base (32.3\% versus 22.3\%) but gains almost nothing from additional turns (32.5\% at \(T10\)), and therefore ends 15.0 points below the base checkpoint at \(T10\). Improved one-shot domain coding thus coexists with a loss of the ability to benefit from iterative feedback. We interpret this as an effect of the continued-pretraining regime itself: causal language modeling on raw repository code, without chat-formatted or instruction data, degrades instruction following and the processing of feedback delivered as conversational turns. The same mechanism is consistent with the structured tool-call degradation reported in Section~\ref{sec:toolrecovery-results}; the interpretation is not tested by a dedicated ablation. SFT on chat-formatted request-to-code pairs, by contrast, improves both the starting point and the final multi-turn outcome.

\begin{figure}[H]
\centering
\includegraphics[width=0.68\textwidth]{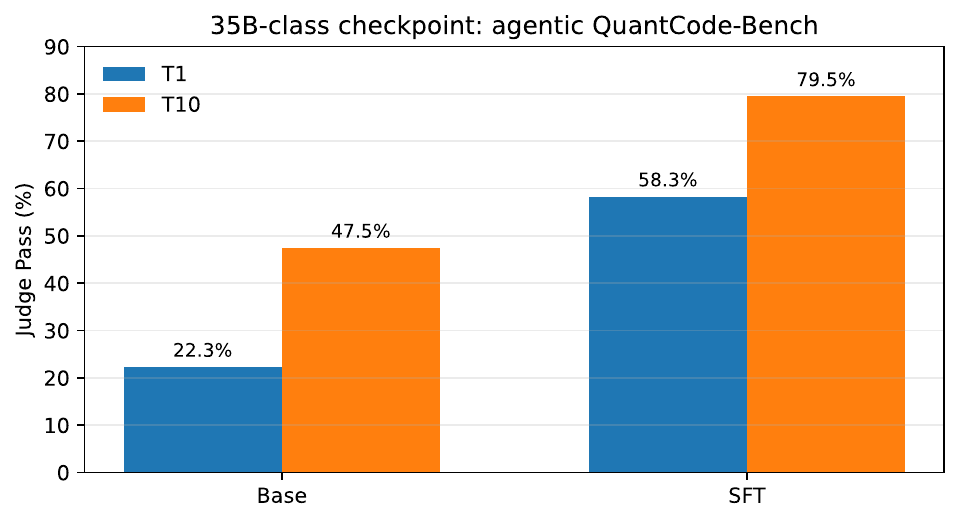}
\caption{Agentic Judge Pass for the Qwen3.6-35B-A3B base and final domain-pretraining-plus-SFT checkpoints. The final checkpoint improves both the initial policy (\(T1\)) and the outcome after up to 10 turns (\(T10\)).}
\label{fig:agentic}
\end{figure}

Figure~\ref{fig:agentic} compares the base and SFT checkpoints directly.

The broader leaderboard shows that iterative feedback benefits general-purpose models as well. Claude Opus 4.6 increases from 75.8\% at \(T1\) to 97.5\% at \(T10\); GPT-5.4 increases from 70.2\% to 95.0\%; Kimi-K2.5 increases from 64.8\% to 93.5\%; and Qwen3-14B increases from 25.2\% to 62.7\%. These measurements reinforce the view that executable strategy generation is naturally agentic: a substantial fraction of failures can be repaired when the model observes the runtime.

\subsection{Domain specialization degrades structured tool-call generation}
\label{sec:toolrecovery-results}

Deployment testing revealed a failure mode that is not directly measured by the preceding benchmark evaluations. After continued pretraining in the causal-language-modeling regime, the Qwen3.6-35B-A3B domain-pretrained checkpoint showed substantial degradation in structured tool-call generation: in deployment checks, tool invocations frequently contained unclosed tool-call tags. The serving engine's parser could not recover a well-formed call from such output, so the invocation was not executed and the agent loop stalled or fell back to plain text. The same defect was observed in the Qwen3.6-35B-A3B strategy-SFT checkpoint, indicating that request-to-code SFT without structured tool-call targets did not restore this formatting behavior. This observation is consistent with a catastrophic-forgetting hypothesis under continued specialization \cite{luo2023forgetting}; more generally, function-calling work emphasizes that successful tool use depends not only on selecting an appropriate API but also on satisfying the structured interface expected by the execution layer \cite{patil2023gorilla}.

The current QuantCode-Bench protocol does not directly measure parser-conformant structured tool calling. Its single-turn protocol extracts a code block from free-form model text, and its agentic protocol supplies structured feedback without requiring the model itself to emit the serving engine's native tool-call syntax. A checkpoint can therefore improve on QuantCode-Bench while regressing on a capability required by a production agent harness. For the intended deployment target---a developer-facing plugin in which the model drives tools directly---this capability is operationally necessary, motivating the recovery experiments of Section~\ref{sec:toolrecovery-method}.

The same regression determines which checkpoints can be evaluated on the repository-level track. Repository tasks are solved through the agent harness, so the domain-pretrained and SFT checkpoints could not be evaluated there: their malformed tool calls stall the agent loop before meaningful repository interaction takes place. Table~\ref{tab:swe} reports this track for general-purpose models. GPT-5.4 solves 38.6\% of the current verified tasks, followed by Qwen3.5-397B-A17B at 33.2\% and Kimi-K2.5 at 29.0\%. Qwen3.5-35B-A3B reaches 25.6\%, while the same model evaluated without reasoning reaches 20.9\%.

\begin{table}[H]
\centering
\caption{Repository-level algorithmic-trading software-engineering track.}
\label{tab:swe}
\begin{tabular}{@{}lr@{}}
\toprule
Model & Tasks solved (\%) \\
\midrule
GPT-5.4 & \textbf{38.6} \\
Qwen3.5-397B-A17B & 33.2 \\
Kimi-K2.5 & 29.0 \\
Qwen3.5-35B-A3B & 25.6 \\
Qwen3.5-35B-A3B, no reasoning & 20.9 \\
\bottomrule
\end{tabular}
\end{table}

The scores are substantially lower than the best \(T10\) values on isolated QuantCode-Bench strategy generation. The task formulations are not numerically comparable, but the lower absolute success rates are consistent with the additional burden of repository navigation, patch construction, multi-file reasoning, test execution, and feedback-driven repair. The 4.7-point gap between reasoning and no-reasoning variants of the Qwen3.5-35B-A3B model further indicates that repository-level algorithmic-trading tasks benefit from explicit multi-step reasoning behavior.

The weight-merge configuration evaluated in this study did not restore parser-conformant tool calling: the 0.65/0.35 linear merge of the SFT checkpoint with the original Qwen3.6-35B-A3B (Section~\ref{sec:toolrecovery-method}) continued to produce malformed tool-call tags in deployment checks, and agentic behavior in the harness was not restored. This result shows that simple post-hoc parameter interpolation was insufficient in our setting, even when more than a third of the weight was placed on the original checkpoint; it does not establish that the capability is unrecoverable by other merging methods or coefficients.

The recovery SFT stage, initialized from this merged checkpoint, produced a different trade-off. Qualitative deployment checks confirmed that both recovery variants could again emit well-formed tool calls accepted by the serving engine: agentic capability that merging alone had failed to recover was partially restored only once the merged checkpoint was additionally trained on agentic trajectories with loss on every assistant turn. Because a standalone parse-success rate was not retained for this snapshot, Table~\ref{tab:toolrecovery} should be interpreted as a measure of downstream agent utility rather than as a quantitative measure of formatting recovery.

\begin{table}[H]
\centering
\caption{Repository-level SWE-bench-like results for the tool-call recovery experiment. Values are percentages of tasks solved, with reasoning disabled. Table~\ref{tab:swe} provides broader model context, but cross-table differences should not be interpreted as effects of the recovery intervention because the tables involve different model generations. The relevant recovery ablation is the within-table comparison against the Qwen3.6-35B-A3B base checkpoint.}
\label{tab:toolrecovery}
\begin{tabular}{@{}lr@{}}
\toprule
Checkpoint & Tasks solved (\%) \\
\midrule
Qwen3.6-35B-A3B base & \textbf{23.7} \\
Qwen3.6-35B-A3B recovery SFT, 16k-token context & 16.3 \\
Qwen3.6-35B-A3B recovery SFT, 32k-token context & 13.0 \\
\bottomrule
\end{tabular}
\end{table}

Although parser-conformant formatting was restored in the qualitative deployment checks, downstream tool-use performance remained below the original base checkpoint. The 16k recovery variant solved 16.3\% of repository tasks compared with 23.7\% for the base, while the 32k variant reached 13.0\%. Thus, off-policy trajectory distillation from Qwen3.5-122B-A10B restored operational compatibility with the tool parser but did not restore the base model's repository-level agent performance. The lower score of the 32k variant is reported as an observation; possible explanations are discussed separately rather than treated as established causes. A further untested factor is the initialization: the recovery SFT was run from the merged checkpoint on the assumption that proximity to the original weights aids recovery, and neither the unmerged SFT checkpoint nor other merge ratios were tried as starting points, so the contribution of the merge to the final result is unknown.

\section{Discussion}

The experiments show that domain specialization is not a single mechanism. Within the directly comparable Qwen3.6-35B-A3B lineage, continued pretraining and subsequent SFT improve different parts of the executable strategy-generation pipeline, while the Qwen3.5-397B-A17B base-to-pretraining comparison independently shows a pretraining gain at larger scale.

Continued pretraining provides the weakest direct supervision but improves domain familiarity. The 5.2- and 6.0-point Judge Pass gains at 35B and 397B scales indicate that additional exposure to algorithmic-trading repositories changes downstream executable behavior. This is important because the intervention does not teach explicit benchmark solutions. Instead, it changes the model's prior over relevant APIs, lifecycle patterns, indicator usage, and code structure. The gain therefore supports the view that framework knowledge is a meaningful bottleneck.

The magnitude of the pretraining gain is nevertheless smaller than the SFT gain. This is consistent with the fact that domain code exposure does not directly solve the semantic translation problem. A model can know Backtrader well and still misunderstand a user's entry condition, reverse an inequality, omit an exit rule, or introduce an indicator not requested by the prompt. SFT reduces this gap by directly pairing strategy intent with executable target code. The validation pipeline further suppresses targets that look reasonable in text but fail in the environment.

The strongest SFT effect appears before any agentic repair. Raising \(T1\) from 22.3\% to 58.3\% means that specialization changes the initial action distribution substantially. This matters operationally: higher first-turn accuracy reduces the expected number of model calls, backtests, and repair iterations required to solve a task. For an interactive coding assistant, this is not merely a benchmark improvement but also a latency and compute improvement.

At the same time, the SFT checkpoint still gains more than 20 percentage points between \(T1\) and \(T10\). Specialization therefore does not eliminate the value of tools. The model remains imperfect at one-shot formalization, but many residual defects are repairable after execution. The behavior of frontier models is similar: large gaps between \(T1\) and \(T10\) are common across the current leaderboard. This suggests that domain-specific code generation benefits from jointly improving the initial generation policy and the repair policy.

The repository-level results expose another boundary. Self-contained strategy generation gives the model a relatively clean problem statement and a known framework. Real software engineering requires locating relevant code in a repository, understanding project-specific abstractions, preserving existing interfaces, modifying multiple files, and interpreting repository tests and their failure output. The low absolute solve rates on the repository track show that this setting remains open even for strong models. It also provides a natural next source of SFT trajectories: real issue resolution, search, patch generation, test execution, and repair inside algorithmic-trading codebases.

The tool-call degradation result carries a broader methodological warning. The primary strategy-generation specialization experiments in this report are evaluated with QuantCode-Bench or closely related execution-based protocols, yet the deployment-critical regression in parser-conformant tool-call syntax was not directly measured by those evaluations and surfaced in the target agent harness. Benchmark-driven specialization should therefore be accompanied by capability-retention checks in the exact serving format, especially when deployment requires structured tool invocation. The recovery experiment also separates two quantities that agentic evaluations can conflate: the probability of emitting a parser-valid call and the utility of the action represented by that call. In our qualitative deployment checks, recovery SFT restored the former, while repository-level performance remained below the base checkpoint.

The difference between the 16k- and 32k-token recovery variants admits several hypotheses, none of which is established by the present experiment. One possible explanation is that the shorter maximum sequence length acts as an implicit filter against long trajectories dominated by repeated failures, exploration loops, or large tool outputs; success-based trajectory filtering has been used in agentic SFT pipelines such as SWE-Gym \cite{pan2024swegym}. A second possibility is token weighting: a small number of long trajectories can contribute a disproportionate fraction of supervised tokens and shift the objective toward late states conditioned on large tool outputs rather than toward high-value decision points. A third hypothesis is off-policy mismatch: later states in teacher-generated trajectories depend on the teacher's preceding actions and may be increasingly unlikely under the student policy. Distinguishing these explanations requires controlled ablations that equalize training-token budgets, filter trajectories by outcome, and compare full versus truncated versions of the same trajectories.

These observations motivate, rather than establish, a dedicated capability-recovery training stage using on-policy trajectory collection. The central point is that restoring syntax is not equivalent to restoring a tool-use policy. A model can again satisfy the parser while remaining worse than the base checkpoint at deciding which tool to call, when to call it, and how to use the result.

Finally, these results reinforce the importance of domain-specific evaluation. General coding benchmarks are necessary but cannot guarantee competence in specialized runtimes. QuantCode-Bench separates compilation, execution, trade generation, and semantic alignment precisely because each stage reveals a different capability. The training experiments demonstrate that this decomposition is also useful diagnostically: continued pretraining and SFT shift different stages by different amounts.

\section{Limitations}

First, the Qwen3.6-35B-A3B base, continued-pretraining, and SFT checkpoints form a sequential lineage evaluated in the current QuantCode-Bench snapshot, and the Qwen3.5-397B-A17B pretraining checkpoint is directly paired with its base. We can therefore make direct stage-by-stage claims for pretraining and SFT within these lineages. However, SFT was applied only in the 35B lineage, so the SFT effect is established at a single scale, and the effect of SFT without prior continued pretraining was not measured.

Second, the internal checkpoints reported in the current QuantCode-Bench snapshot were produced during an evolving training program. The present report describes the data sources, validation logic, and evaluation protocol that determine the conceptual method, but it does not claim a complete public reproduction package for every historical run. Future releases should freeze dataset versions, framework versions, optimizer configurations, random seeds, and exact training schedules.

Third, the LLM judge is an automated semantic proxy rather than an absolute oracle. A judge can miss subtle trading-logic mismatches or exhibit model-specific biases. QuantCode-Bench mitigates this by placing judging after executable checks, but semantic evaluation remains a source of uncertainty \cite{khoroshilov2026quantcode}.

Fourth, neither QuantCode-Bench nor the SFT validation pipeline measures profitability, robustness across market regimes, realistic transaction costs, market impact, or production risk. The task is faithful implementation of requested code. The reported success rates must not be interpreted as evidence that the generated strategies are profitable or suitable for deployment.

Fifth, continued pretraining on public repository code introduces the standard contamination concern for code benchmarks. The benchmark and training data are constructed from overlapping broad domains, and future work should strengthen repository-level decontamination and exact-match filtering. The repository track in particular should be versioned so that issue/patch leakage can be audited.

Sixth, the tool-call degradation and recovery observations were identified in deployment-oriented checks, but a standalone parser-validity rate was not retained for the internal snapshot used here. We therefore report formatting recovery qualitatively and repository task resolution quantitatively. Future evaluations should instrument exact tool-call parse success, argument validity, and downstream task success as separate metrics.

Finally, the current repository-level track is approximately 200 verified tasks and remains under active development. It is large enough to expose substantial differences among models, but a broader set of repositories, frameworks, languages, and market-data environments would be needed to characterize general algorithmic-trading software-engineering ability.

\section{Conclusion}

Domain-specific training materially improves executable algorithmic-trading code generation. In the sequential Qwen3.6-35B-A3B lineage, continued pretraining raises single-turn Judge Pass from 27.8\% to 33.0\%, and subsequent agent-validated SFT raises it further to 58.2\%. The final 35B checkpoint also reaches 58.3\% first-turn agentic success and 79.5\% success after up to 10 turns, compared with 22.3\% and 47.5\% for the original base checkpoint. At 397B scale, continued pretraining directly improves Judge Pass from 41.5\% to 47.5\%.

We also identify a capability-retention failure outside the main strategy-generation benchmark. Domain specialization degraded parser-conformant structured tool calling in the target agent harness. Dedicated recovery SFT restored operational tool-call formatting in qualitative deployment checks, but repository-level task resolution remained below the Qwen3.6-35B-A3B base checkpoint. This result separates interface compatibility from effective tool-use policy and shows that specialization should be evaluated against both target-domain gains and retained agent capabilities.

The central conclusion is that specialized code generation should be trained and evaluated as behavior in an environment, not as text alone. Framework-oriented pretraining improves the model's prior knowledge of the software ecosystem. Validated SFT teaches a direct mapping from user intent to idiomatic executable code. Capability-retention checks are additionally required when the deployed system depends on structured tool use.

The next step is to repeat the full pretraining-and-SFT sequence at additional model scales under the same frozen benchmark version. A second direction is to train on repository-level trajectories involving code search, issue interpretation, patch generation, repository tests, and multi-turn repair. A third is to instrument structured tool calling as a first-class retained capability and evaluate parser validity separately from downstream tool-use quality. A fourth is to incorporate real user trajectories so that the distribution of requests and failure corrections more closely reflects practical algorithmic-trading development. Together, these directions can turn domain-specific code models from one-shot generators into robust software agents that operate reliably inside specialized execution environments.

\end{document}